\documentclass{article}
\usepackage[preprint,nonatbib]{nips_2018}

\usepackage[utf8]{inputenc} % allow utf-8 input
\usepackage[T1]{fontenc}    % use 8-bit T1 fonts
\usepackage{hyperref}       % hyperlinks
\usepackage{url}            % simple URL typesetting
\usepackage{booktabs}       % professional-quality tables
\usepackage{amsfonts}       % blackboard math symbols
\usepackage{nicefrac}       % compact symbols for 1/2, etc.
\usepackage{microtype}      % microtypography
\usepackage{amsmath} 
\usepackage{color,soul}
\usepackage{listings}
\usepackage{mathtools}

\newcommand{\mat}[1]{{\bf #1}}

\definecolor{codegreen}{rgb}{0,0.6,0}
\definecolor{codegray}{rgb}{0.5,0.5,0.5}
\definecolor{codepurple}{rgb}{0.58,0,0.82}
\definecolor{backcolour}{rgb}{0.95,0.95,0.92}

\lstdefinestyle{mystyle}{
  backgroundcolor=\color{backcolour},   
  commentstyle=\color{codegreen},
  keywordstyle=\color{magenta},
  numberstyle=\tiny\color{codegray},
  stringstyle=\color{codepurple},
  basicstyle=\footnotesize,
  breakatwhitespace=false,         
  breaklines=true,                 
  captionpos=b,                    
  keepspaces=true,                 
  numbers=left,                    
  numbersep=5pt,                  
  showspaces=false,                
  showstringspaces=false,
  showtabs=false,                  
  tabsize=2
}

\title{McTorch, a manifold optimization\\ library for deep learning}

\author{
Mayank Meghwanshi\\
Microsoft, India\\
\texttt{mamegh@microsoft.com}
\And
Pratik Jawanpuria\\
Microsoft, India\\
\texttt{pratik.jawanpuria@microsoft.com}
\And
Anoop Kunchukuttan\\
Microsoft, India\\
\texttt{anoop.kunchukuttan@microsoft.com}
\And
Hiroyuki Kasai\\
The University of Electro-Communications, Japan\\
\texttt{kasai@is.uec.ac.jp}
\And
Bamdev Mishra\\
Microsoft, India\\
\texttt{bamdevm@microsoft.com}
}
\begin{document}
% \nipsfinalcopy is no longer used

\maketitle

\begin{abstract}
In this paper, we introduce McTorch, a manifold optimization library for deep learning that extends PyTorch\footnote{PyTorch is available at \url{https://pytorch.org/} and introduced in \cite{paszke17a}.}. It aims to lower the barrier for users wishing to use manifold constraints in deep learning applications, i.e., when the parameters are constrained to lie on a manifold. Such constraints include the popular orthogonality and rank constraints, and have been recently used in a number of applications in deep learning. McTorch follows PyTorch's architecture and decouples manifold definitions and optimizers, i.e., once a new manifold is added it can be used with any existing optimizer and vice-versa. McTorch is available at \url{https://github.com/mctorch}.
\end{abstract}

\section{Introduction}
Manifold optimization refers to nonlinear optimization problems of the form
\begin{equation}\label{eq:manifold_optimization}
    \min_{x \in \mathcal{M}}\  f(x),
\end{equation}
where $f$ is the loss function and the parameter search space $\mathcal{M}$ is a smooth \emph{Riemannian} manifold \cite{absil08a}. Note that the Euclidean space is trivially a manifold. Conceptually, manifold optimization translates the constrained optimization problem (\ref{eq:manifold_optimization}) into an unconstrained optimization over the manifold $\mathcal{M}$, thereby generalizing many of the standard nonlinear optimization algorithms with guarantees \cite{absil08a,bonnabel13a,sato13a,Sato17a,zhang16a,kasai18a,boumal2018globalrates}. A few ingredients of manifold optimization are the matrix representations of the tangent space (linearization of the search space at a point), an inner product (to define a metric structure to compute the Riemannian gradient and Hessian efficiently), and a notion of a straight line (a characterization of the geodesic curves to maintain strict feasibility of the parameters). Two popular examples\footnote{A comprehensive list of manifolds is at \url{http://manopt.org/tutorial.html}.} of smooth manifolds are i) the {Stiefel} manifold, which is the set of $n\times p$ matrices whose columns are orthonormal, i.e., ${\mathcal{M} \coloneqq \{ \mat{X} \in \mathbb{R}^{n\times p} : \mat{X}^\top\mat{X} = \mat{I} \}}$ \cite{edelman98a} and ii) the symmetric positive definite manifold, which is the set of symmetric positive definite matrices, i.e., ${\mathcal{M} \coloneqq \{ \mat{X} \in \mathbb{R}^{n\times n} : \mat{X}\succ \mat{0} \}}$ \cite{bhatia09a}.

Manifold optimization has gained significant interest in computer vision \cite{kovnatsky2016madmm, tron2017space}, Gaussian mixture models \cite{hosseini2015matrix}, multilingual embeddings \cite{jawanpuria18a}, matrix/tensor completion \cite{boumal14a,mjaw18a,Kasai2016a,kressner13a,mishra14a,nimishakavi18a,vandereycken13a}, metric learning \cite{meyer2011linear,zadeh2016geometric}, phase synchronization \cite{boumal2016nonconvexphase,zhong2018nearoptimal}, to name a few. 

Deep learning refers to machine learning methods with multiple layers of processing to learn effective representation of data. These methods have led to state-of-the-art results in computer vision, speech, and natural language processing. Recently, manifold optimization has been applied successfully in various deep learning applications \cite{nickel18a,arjovsky16a,roy18a,huang17a,huang18a,ozay18a,badrinarayanan15a}. 

On the practical implementation front, there exists popular toolboxes -- Manopt \cite{boumal14a}, Pymanopt \cite{townsend16a}, and ROPTLIB  \cite{huang16a} -- that allow rapid prototyping without the burden of being well-versed with manifold-related notions. However, these toolboxes are more suitable for handling standard nonlinear optimization problems and not particularly well-suited for deep learning applications. On the other hand, PyTorch \cite{paszke17a}, a Python based deep learning library, supports tensor computations on GPU and provides dynamic tape-based auto-grad system to create neural networks. PyTorch provides a flexible format to define and train deep learning networks. Currently, however, PyTorch lacks manifold optimization support. The proposed McTorch\footnote{McTorch stands for \textbf{M}anifold-\textbf{c}onstrained \textbf{Torch}.} library aims to bridge this gap between the standard manifold toolboxes and PyTorch by extending the latter's functionality. 

McTorch builds upon the PyTorch library  for tensor computation and GPU acceleration, and derives manifold definitions and optimization methods from the toolboxes \cite{boumal14a,huang16a,townsend16a}. McTorch is well-integrated with PyTorch that allows users to use manifold optimization in a straightforward way. 

\section{Overview of McTorch}\label{sec:overview}

McTorch library has been implemented by extending a PyTorch fork to closely follow its architecture. All manifold definitions reside in the module \texttt{torch.nn.manifold} and are derived from the parent class \texttt{Manifold}, which defines the manifold structure (i.e., the expressions of manifold-related notions) that any manifold must implement. A few of these expressions are: 
\begin{itemize}\setlength\itemsep{0em}
\item \texttt{rand}: to get a random point on the manifold, 
\item \texttt{retr}: to retract a tangent vector onto the manifold, 
\item \texttt{egrad2rgrad}: to convert the back-propagated gradient to the Riemannian gradient.
\end{itemize}

To facilitate creation of a manifold-constrained parameter, PyTorch's native 
\texttt{Parameter} class is modified to accept an extra argument on initialization to specify the manifold type and size. \texttt{Parameter} can be initialized to a random point on the manifold or a particular value provided by the user. \texttt{Parameter} also holds the attribute \texttt{rgrad} (which stands for the Riemannian gradient) that gets updated with every back-propagated gradient step.

The existing optimizers in the module \texttt{torch.optim} are modified to support updates on the manifold. An optimization step is a function of parameter's current value, gradient, and optimizer state. In the manifold optimization, the gradient is the Riemannian gradient and the update is with the retraction operation.

To use manifolds in PyTorch layers (in \texttt{torch.nn.Module}), we have added the property \texttt{weight\_manifold} to the linear and convolutional layers which constrains the weight tensor of the layer to a specified manifold. As the shape of weight tensor is calculated using the inputs to the layer, we have added \texttt{ManifoldShapeFactory} to create a manifold object for a given tensor shape such that it obeys the initialization conditions of that manifold.

All the numerical methods are implemented using the tensor functions of PyTorch and support both CPU and GPU computations. As the implementation modifies and appends to the PyTorch code, all the user facing APIs are similar to PyTorch. McTorch currently supports:
\begin{itemize}\setlength\itemsep{0em}
\item \textbf{Manifolds}: \texttt{Stiefel}, \texttt{PositiveDefinite},
\item \textbf{Optimizers}: \texttt{SGD}, \texttt{Adagrad}, \texttt{ConjugateGradient},
\item \textbf{Layers}: \texttt{Linear}, \texttt{Conv1d}, \texttt{Conv2d}, \texttt{Conv3d}. 
\end{itemize}

\section{McTorch Usage} \label{sec:using_mctorch}
%McTorch can be used for manifold optimization on tensors and deep learning modules.

%\textbf{Principal components analysis.} McTorch for optimization can be setup with just three simple steps. Below example shows PCA computation as an optimization problem on Stiefel manifold,
 %
%\begin{lstlisting}[language=python]
%import torch
%import torch.nn as nn
%
%# Random data with high variance in first two dimension
%X = torch.diag(torch.FloatTensor([3,2,1])).matmul(torch.randn(3,200))
%
%# 1. Initialize Parameter
%manifold_param = nn.Parameter(manifold=nn.Stiefel(3,2))
%
%# 2. Define Cost - squared reconstruction error
%def cost(X, w):
%    wTX = torch.matmul(w.transpose(1,0), X)
%    wwTX = torch.matmul(w, wTX)
%    return torch.sum((X - wwTX)**2)
%
%# 3. Optimize
%optimizer = torch.optim.Adagrad(params = [manifold_param], lr=0.5)
%for epoch in range(30):
%    cost_step = cost(X, manifold_param)
%    print(cost_step)
%    cost_step.backward()
%    optimizer.step()
%    optimizer.zero_grad()
%\end{lstlisting}

\textbf{Orthogonal weight normalization in multi-layer perceptron} \cite{huang17a}. An example showing creation and optimization of McTorch module with the Stiefel manifold.

\begin{lstlisting}[language=python]
import torch
import torch.nn as nn
import torch.nn.functional as F

# A McTorch module using manifold constrained linear layers.
class OrthogonalWeightNormalizationNet(nn.Module):
    def __init__(self, input_size, hidden_sizes, output_size):
        super(OrthogonalWeightNormalizationNet, self).__init__()
        layer_sizes = [input_size] + hidden_sizes + [output_size]
        self.layers = []
        
        for i in range(1, len(layer_sizes)):
            self.layers.append(nn.Linear(in_features=layer_sizes[i-1], 
                                        out_features=layer_sizes[i], 
                                        weight_manifold=nn.Stiefel))
            # ReLU for middle layers
            if i != len(layer_sizes)-1:
                self.layers.append(nn.ReLU())
            # LogSoftmax at the output layer
            else:
                self.layers.append(nn.LogSoftmax(dim=1))
        self.model = nn.Sequential(*self.layers)

    def forward(self, x):
        return self.model(x)

# Create module object.
model = OrthogonalWeightNormalizationNet(input_size=1024,
        hidden_sizes=[128, 128, 128, 128, 128], output_size=68)

# Optimize with the Adagrad algorithm.
optimizer = torch.optim.Adagrad(params=model.parameters(), lr=1e-2)
for epoch in range(10):
    optmizer.zero_grad()
    data, target = get_next_batch()
    output = model(data)
    loss = F.nll_loss(output, target)
    cost.backward()
    optimizer.step()
\end{lstlisting}

In the above example, a new PyTorch module is defined by inheriting from \texttt{nn.Module}. It requires defining an \texttt{\_\_init\_\_} function to set up layers and a \texttt{forward} function to define forward pass of the module. The backward pass for back-propagation of gradients is automatically computed. In the layer definition, \texttt{Linear} layers with \texttt{Stiefel} (orthogonal) manifold are initialized. It also adds ReLU nonlinearity between layers and softmax nonlinearity on the output. The forward function of the model does a forward pass on sequence of layers. To optimize, \texttt{torch.optim.Adagrad} is initialized, which is followed by multiple epochs of forward and backward passes of the module on batched training data.

\section{Roadmap and Conclusion}\label{sec:conclusion}
We are actively working on adding support for more manifolds and optimization algorithms. We are also curating a collection of code examples and benchmarks to showcase various uses of manifold optimization in deep learning research and applications. 

McTorch is released under the BSD-3 Clause license and all the codes and examples are available on the GitHub repository of the project at \url{https://github.com/mctorch/mctorch}.

\bibliography{mctorch2018}
\bibliographystyle{plain}

\end{document}